\documentclass[10pt, conference, a4paper]{IEEEtran}

\usepackage{cite}

\ifCLASSINFOpdf
   \usepackage[pdftex]{graphicx}
\else
\fi

\usepackage{amsmath}
\usepackage{amsfonts}
\usepackage{gensymb}
\DeclareMathOperator*{\argmax}{arg\,max}

\usepackage[outdir=./]{epstopdf}

\usepackage{svg}

\ifCLASSOPTIONcompsoc
  \usepackage[caption=false,font=normalsize,labelfont=sf,textfont=sf]{subfig}
\else
\usepackage{subfig}
\fi

\usepackage{hyperref}

\begin{document}
\title{Deep Ordinal Regression with Label Diversity}

\author{\IEEEauthorblockN{Axel Berg\textsuperscript{1,2}, Magnus Oskarsson\textsuperscript{2}, Mark O'Connor\textsuperscript{1}}
\IEEEauthorblockA{\textsuperscript{1}Arm Research \\
Email: \{axel.berg,mark.oconnor\}@arm.com \\
\textsuperscript{2}Center for Mathematical Sciences, Lund University, Sweden\\
Email: \{magnus.oskarsson\}@math.lth.se}}

\maketitle

\begin{abstract}
Regression via classification (RvC) is a common method used for regression problems in deep learning, where the target variable belongs to a set of continuous values. By discretizing the target into a set of non-overlapping classes, it has been shown that training a classifier can improve neural network accuracy compared to using a standard regression approach. However, it is not clear how the set of discrete classes should be chosen and how it affects the overall solution. In this work, we propose that using several discrete data representations simultaneously can improve neural network learning compared to a single representation. Our approach is end-to-end differentiable and can be added as a simple extension to conventional learning methods, such as deep neural networks. We test our method on three challenging tasks and show that our method reduces the prediction error compared to a baseline RvC approach while maintaining a similar model complexity.
\end{abstract}

\IEEEpeerreviewmaketitle

\section{Introduction}

Choosing the right objective function is a crucial part of successfully training an accurate and generalized regression model, for example a deep neural network. Among the standard objective functions are the mean squared error (MSE, or $L^2$ loss), the mean absolute error (MAE, or $L^1$-loss) and hybrid variants such as the Huber loss. Much attention has also been given to deriving problem-specific objective functions that incorporate certain aspects of the target variable, such as modularity and norm constraints for geometric regression. It is also possible to treat a continuous dependent variable as belonging to a finite number of discrete classes, although this necessarily comes at the expense of introducing a discretization error. Such approaches are known as regression via classification (RvC) and they are frequently used in tasks where a regression loss would at first seem more natural \cite{workman2016horizon, wang2015designing, zeisl2014discriminatively, tulsiani2015viewpoints, abbas2019geometric, rothe2015dex}. 

Ordinal regression techniques can be applied to classification problems where the dependent variable exhibits a relative ordering. Techniques for ordinal regression can be applied to RvC problems in order to preserve the ordinal structure of the labels, and recent work has shown that this method can be used to improve the accuracy in several regression problems, such as age estimation \cite{niu2016ordinal} \cite{beckham2017unimodal}, image ranking and depth estimation \cite{diaz2019soft}.

One problem with the RvC approach is the ambiguity in how the discrete classes should be created from the distribution of the dependent variable. The standard approach is to create bins of equal width covering the target output range. For skewed distributions one can also apply the method of equal frequency, where the bins are created from the cumulative distribution function of the target, such that each bin contains the same number of training examples. Regardless of the method, the number of bins must be selected and optimized for the given task, which raises the question of what the optimal number of bins is for a given problem. If the bin-width is too small, this can result in few training examples in each class, but if it is too large, the discretization error can become a limiting factor.

The ambiguity in how to bin a continuous variable leads to a diverse range of possibilities in how to represent the target values --- a fact that can be exploited. From previous research it is well known that a diverse ensemble of individual predictors can be combined in order to reduce the overall prediction error. Such approaches often involves training multiple regressors where the diversity is ensured by either data augmentation or model selection. This leads to increased overhead at both training and inference time \cite{ren2016ensemble}. However, with the use of label diversity and deep neural networks, one can create a multi-output predictor that enforces diversity without extra computational complexity.

\textbf{Contribution:} In this paper, we show that a collection of different binning variants of the target values can be used to improve prediction accuracy without increasing the computational complexity compared to a standard classifier. We do so by training a deep multi-output convolutional neural network (CNN) to classify training examples in multiple overlapping bins simultaneously. By doing so, we can effectively take into account the ordinal structure of the regression problem, while also making use of the diversity of the different possible representations of the target variable. We demonstrate our method on a number of different tasks and show competitive results compared to current state-of-the-art methods. 

\begin{figure}[t]
  \centering
  \includegraphics[width=\linewidth]{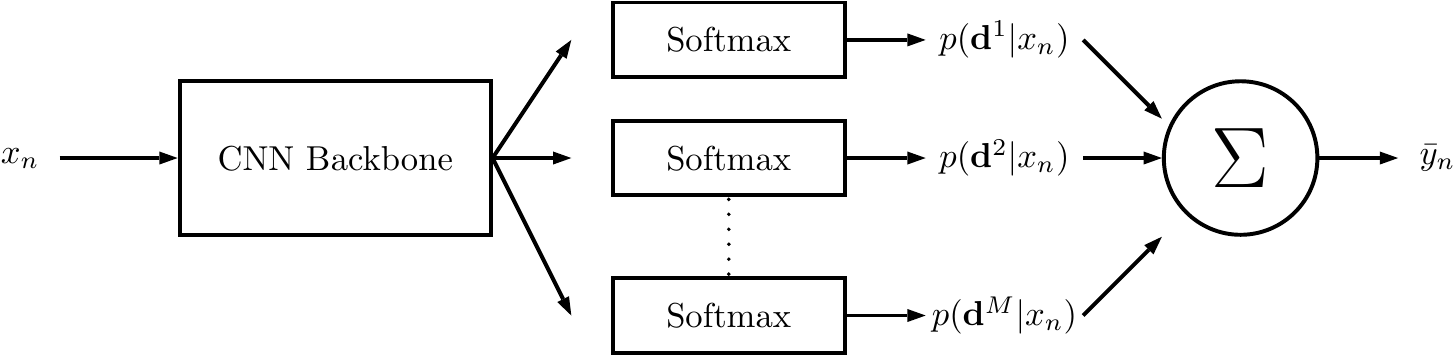}
  \caption{An illustration of the neural network architecture with multiple output heads. At inference time the expected values of the different distributions are combined using an ensemble average.}
\label{inference}
\end{figure}

\section{Related Work}

\subsection{Methods}

\subsubsection{Ordinal Regression}
Ordinal regression, or ranking learning, is used for problems where the target variable exhibits a relative ordering on an arbitrary scale, e.g. categories such as "bad", "good" and "very good". When performing vanilla RvC, the ordinal information contained in the target values is lost, but this can be resolved by using ordinal regression techniques. A common variant is to use extended binary classification, where a single multi-class classification problem is reduced to a set of several binary classification problems \cite{li2007ordinal}. With the advance of deep learning in recent years, ordinal regression has been used successfully for several tasks, including monocular depth estimation \cite{fu2018deep}, age estimation \cite{niu2016ordinal}, head pose estimation \cite{hsu2018quatnet}, medical diagnosis \cite{liu2018ordinal} and historical image dating \cite{liu2017deep}.

\subsubsection{Ensemble Learning}
The fact that a set of individual regressors or classifiers can be combined into an ensemble in order to reduce the overall prediction accuracy of a model is frequently exploited in machine learning research \cite{ren2016ensemble}. A key notion in ensemble learning is the bias-variance-covariance decomposition, which says that in order for an ensemble to reduce the prediction error, there has to be some variance in the predictions of the ensemble members. Therefore, the aim should be to create ensembles that consist of accurate but diverse predictors.

Diversity can be created in several ways, most commonly through methods like bagging \cite{breiman1996bagging} and boosting \cite{freund1996experiments}, which rely on different forms of data diversity. However, using such method comes at the cost of extra model complexity and training time. Recently, Zhang et al.  \cite{zhang2019robust} proposed a framework for training an ensemble of networks using negative correlation learning, where the high level features are learned via parameter sharing. This reduces the overhead, while keeping the benefits of a diverse ensemble. In general, a multi-output neural network can be used to form an ensemble, which we exploit in our proposed method.

\subsection{Relevant Applications}
\subsubsection{Age Estimation}
Deep learning methods have been used successfully for age estimation, where the task consists of predicting the age of a person given a single RGB image of the person's face. Depending on the implementation, a person's age can be considered either as a continuous variable on the positive real numbers, or as a class belonging to a set of discrete positive integers. Rothe et al. \cite{rothe2015dex} first highlighted the use of end-to-end training of CNNs for age estimation from a single image. The authors noted that classification yielded better results than direct regression and since then, several new methods using ordinal regression techniques have been published. 

Agustsson et al. \cite{agustsson2017anchored} performed end-to-end piecewise linear regression by assigning each regressor to an anchor point. Others have observed that it is easier for a human to distinguish differences in age between two persons, rather than their absolute age and used this as a design principle for ordinal regression \cite{zhang2017quantifying, niu2016ordinal}. Alternative methods have focused on various soft encodings of the age over classes, where the elements of the probability vector are proportional to the distance from the true class \cite{gao2017deep, zeng2019soft, diaz2019soft}. In this way both the ordinal and metric information can be effectively encoded in the labels. Furthermore, it has also been shown that forcing the output of the classifier to be rank consistent over ages can improve the overall accuracy \cite{beckham2017unimodal, cao2019rank}.

\subsubsection{Head Pose Estimation}
Head pose estimation is the task of predicting the pose of a human head with three degrees of freedom, given an image and possibly depth information \cite{murphy2008head}. There are several ways to represent the pose, including three rotation angles (pitch, yaw and roll) with respect to a set of principal axes, a $3\times 3$ rotation matrix or a single quaternion. During the past years, several CNN-based head pose estimators trained on specific loss functions have been proposed. Chang et al. \cite{chang2017faceposenet} combined direct head pose regression with facial landmark detection and used it for facial alignment. Ruiz et al. \cite{ruiz2018fine} used a multi-loss CNN and showed that using a balanced hybrid variant of regression and classification yielded improvements over previous methods. Ordinal methods have also been used, such as in \cite{hsu2018quatnet}, where a combined ranking and MSE regression loss is used in conjunction with a quaternion representation of the head pose. The results indicated that training the CNN to regress the three angles while simultaneously solving several binary ranking problems improved the prediction accuracy compared to a standard regression or classification baseline. 

\subsubsection{Historical Image Dating}
Palermo et al. \cite{DBLP:conf/eccv/PalermoHE12} first introduced the task of automatically estimating the historical time in which a color photograph was taken using machine learning techniques. The authors noted that there are several features of the color imaging process that are typical to the era in which the images were taken, such as hue, saturation and color histogram. They then used a support vector machine to classify the images into different decades and showed that this method vastly outperformed untrained humans in terms of classification accuracy. Ginosar et al. \cite{ginosar2015century} used American high school yearbooks to train a deep neural network for the same task, but in this case the extracted features were also dependent on the image content, e.g. facial attributes and hairstyle. Recently, ordinal regression techniques have also been applied successfully to the task of image dating \cite{liu2017deep, belharbi2019deep}. 

\section{Proposed Method}

\begin{figure}[t]
  \centering

  \includegraphics[width=\linewidth]{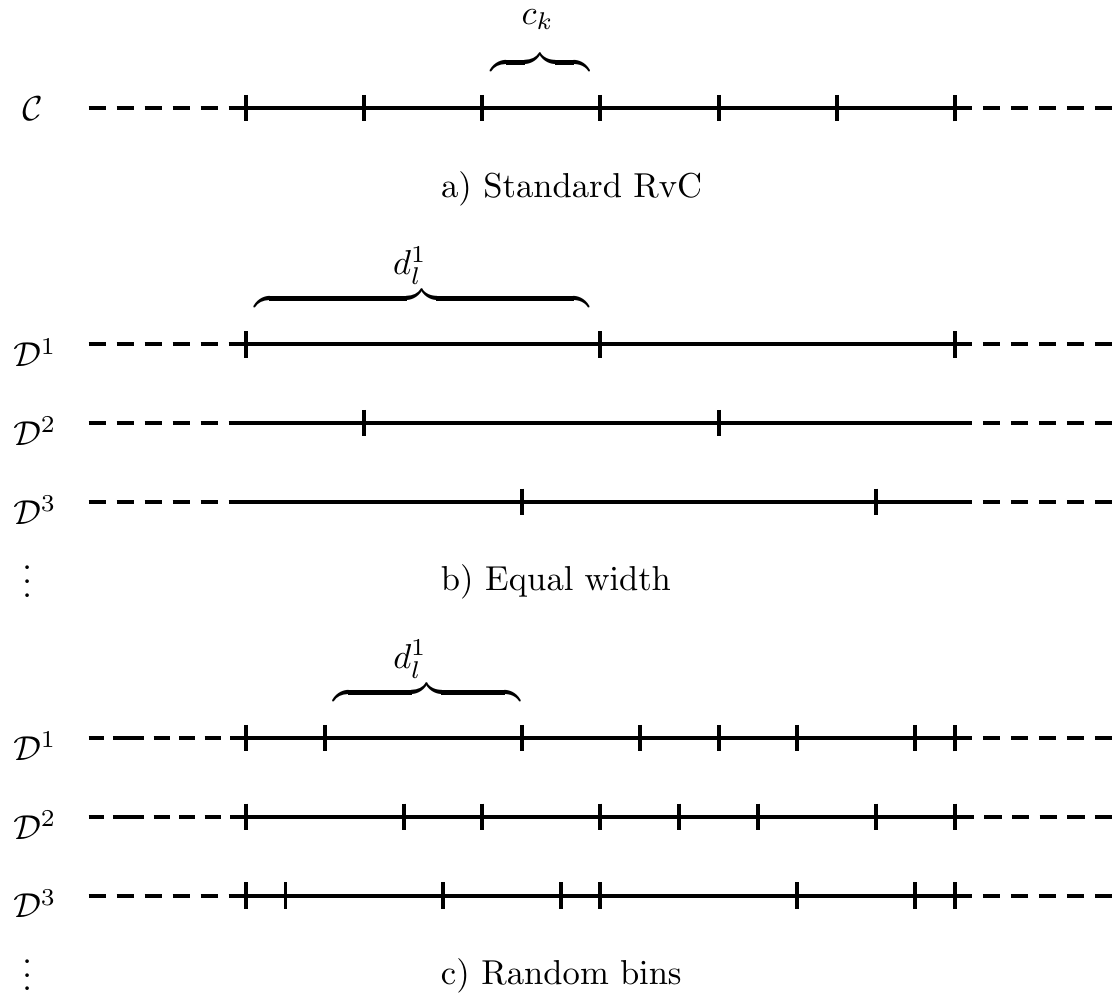}
  \caption{Examples of how the sets of class intervals $\mathcal{D}^m$ can be constructed using the different methods.}
 \label{classes_fig}
\end{figure}

\subsection{Label Diversity by Overlapping Bins}
Let $x_n \in \mathbb{R}^p$ denote the $n^\text{th}$ input (independent variable) and let $t_n \in \mathbb{R}$ be the corresponding target value (dependent value) for $n = 1, ..., N$. Now let $\mathcal{C} = \{c_k\}_{k=1}^{K}$ be a set of of non-overlapping intervals on the real line $\mathbb{R}$, as shown in Figure \ref{classes_fig}a, such that $\cup_{k=1}^{K} c_k$ covers the samples $t_n$ for all $n$. The standard RvC approach would now be to map each target value $t_n$ to a unique class $c_k$ and train a classifier to predict the posterior probability over classes $p(c_k | x_n)$.

In order to create label diversity, we instead introduce $M$ new sets of class intervals $\mathcal{D}^m = \{d^m_l\}_{l=1}^{L_m}$ for $m = 1,..., M$, such that $\cup_{l=1}^{L_m} d_l^m =  \cup_{k=1}^{K} c_k$, $\forall m$. By doing so, we have created several new discretizations that all cover the support of the target value, but in different ways. Here we do not provide an answer to exactly how these discretizations ought to be chosen, since this in itself is an optimization problem where the solution most likely depends on the problem domain, but instead we focus on the two following possibilities, where we assume that $L_m = L$, $\forall m$, such that each discretization contains equally many classes:
\begin{enumerate}

\item Assuming that we do not want to increase the complexity of the training algorithm compared to the standard RvC approach, we fix the total number of classes such that $ML = K$. Then we create $M$ discretizations, each containing $L$ equally wide bins, such that the overlap between $d_l^m$ and $d_l^{m+1}$ is fixed. An illustration of this approach is shown in Figure \ref{classes_fig}b. We refer to this method as "equal width".
\item In order to maximize diversity between different discretizations, for each $m = 1, ..., M$, we randomly sample $L < K$ classes (with replacement) from $\mathcal{C}$ and let these classes be the centers of the new bins in $\mathcal{D}^m$, such that target values that do not belong to any of the chosen classes are assigned to the nearest neighbor in the sample. An illustration of this approach is shown in Figure \ref{classes_fig}c We refer to this method as "randomized bins".
\end{enumerate}

If the target is multi-dimensional, the same methods can be applied by creating classes for each dimension individually. 

\subsection{Backpropagation}
\begin{figure}[t]
  \centering
  \includegraphics[width=\linewidth]{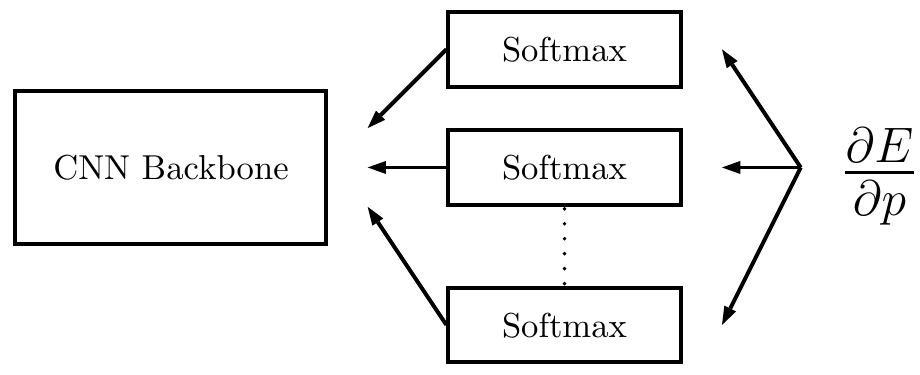}
  \caption{During training, the loss is backpropagated through the different softmax heads to the previous layers.}
\label{backprop}
\end{figure}

The proposed methods can be used together with a neural network by replacing the last fully connected layer and activation with $M$ fully connected layers of size $L$ in parallel, where each layer has its own softmax activation. The network is trained by minimizing the sum of the negative cross-entropies between the individual classifier and the targets over each mini-batch of size $N_b$. The loss function then becomes
\begin{equation}
E = - \sum_{n = 1}^{N_b} \sum_{m=1}^M \sum_{l=1}^{L_m} q_n(d_l^m) \log p(d_l^m |x_n) ,
\label{ce}
\end{equation}
where $q_n(d_l^m)$ is a one-hot encoding of the binned target value, such that the only non-zero element is the one where the bin overlaps the true target:
\begin{equation}
q_n(d_l^m) = \begin{cases}
1, \ t_n \in d_l^m \\ 
0, \quad \text{otherwise}
\end{cases}
\end{equation}
The predictions $p(d_l^m |x_n)$ are computed using the $M$ individual softmax heads of size $L$, and the loss is then back-propagated through the network by differentiating with respect to the predictions as shown in Figure \ref{backprop}.

In order to see how the loss function incorporates the ordinal relationship between the targets, we can again consider Figure \ref{classes_fig}. In order to make a correct prediction, each softmax head must output a high probability for the correct class in each discretization. If the output probabilities are correct for only a subset of the $M$ different discretizations (which implies that the prediction is slightly off) then this will be penalized by an increased loss.

\subsection{Inference}
At inference time, the output posterior should be evaluated and converted to a point estimate of the target value by a hard decision. For standard RvC methods, this is typically done by either taking the expected value of the output distribution over classes or using the maximum-a-posteriori estimator. Therefore, we propose two similar methods to perform inference.

For problems where the target value $t_n$ belongs to the real line, we do this by computing the expected value over each estimated posterior distribution as
\begin{equation}
\hat{y}_{m,n} = \sum_{l=1}^{L_m} w^m_l p(d^m_l | x_n),
\label{expected}
\end{equation}
where $w^m_l$ denotes the mean value of the bin $d^m_l$. This gives us $M$ different estimates of the target, which are then combined using a weighted average. Here we only consider uniform weighting of the individual estimates, i.e.
\begin{align}
\bar{y}_n = \frac{1}{M} \sum_{m=1}^M \hat{y}_{m,n}.
\label{ybar}
\end{align}
This is a form of ensemble average, and we note that we can decompose an individual error made by the ensemble using the ambiguity decomposition \cite{brown2003use} as
\begin{equation}
(\bar{y}_n - t_n)^2 = \frac{1}{M} \sum_{m=1}^M (\hat{y}_{n, m} - t_n)^2 - \frac{1}{M} \sum_{m=1}^M (\hat{y}_{n, m} - \bar{y}_n)^2 ,
\label{amb}
\end{equation}
which shows that the ensemble error is less than the average error of the individual estimates if there is enough variance within the ensemble. In this case, the non-zero variance is guaranteed by the different weights associated with each expected value in equation (\ref{expected}).

On the other hand, if the target belongs to a set of ordinal classes where we cannot define a useful distance metric, it is more suitable to use the MAP estimate. We do this by marginalizing over $d^m_l$ in order to estimate the average posterior over the original classes $c_k$ as

\begin{equation}
p(c_k | x_n) = \frac{1}{M} \sum_{m=1}^M \sum_{l=1}^{L_m} p(c_k | d^m_l, x_n) p(d^m_l | x_n),
\end{equation}
where we assume that the conditional probability $p(c_k | d^m_l, x_n)$ is independent of the input $x_n$, i.e.
\begin{equation}
p(c_k | d^m_l, x_n) = \frac{||d^m_l \cap c_k||}{||d^m_l||} .
\end{equation}
Then we compute the MAP estimate as 
\begin{equation}
k^* = \argmax_k p(c_k | x_n) .
\label{map}
\end{equation}
This shows that our method can be used both for true regression problems where we can measure distances between target values and for classification problems where an ordering of targets exists, but without a well-defined distance. In section V we show experiments for tasks of both the first and second kind.

\section{An Illustrative Example}
In order to demonstrate the ensemble-like effect of our method, we train a shallow CNN on the task of predicting the rotation angle of digits from the MNIST dataset of handwritten digits \cite{lecun1998mnist}. The dataset consists of 5,000 training images and 5,000 test images, where a digit is rotated by an integer drawn uniformly in the interval $t_n \in [-45 \degree, 45 \degree]$. We implemented label diversity using the randomized bins method for all combinations of $M \in [2, 4, 16, 8, 32, 64]$ and $L \in [8, 16, 32, 64]$. We then trained a five-layer CNN, with $M$ softmax heads, each containing $L$ output units, to predict the one-hot encoded labels of the rotation angles. At inference time, we used equation (\ref{ybar}) for prediction and evaluated the MAE on the test set. For comparison, we also trained a regression baseline by using the same shallow CNN with the MSE loss, but where the softmax heads were replaced by a single output unit with a linear activation. The training process was repeated for 10 random initializations of the network and the MAE was averaged over the 10 different trials.

\begin{figure}[t]
  \centering
  \includegraphics[width=\linewidth]{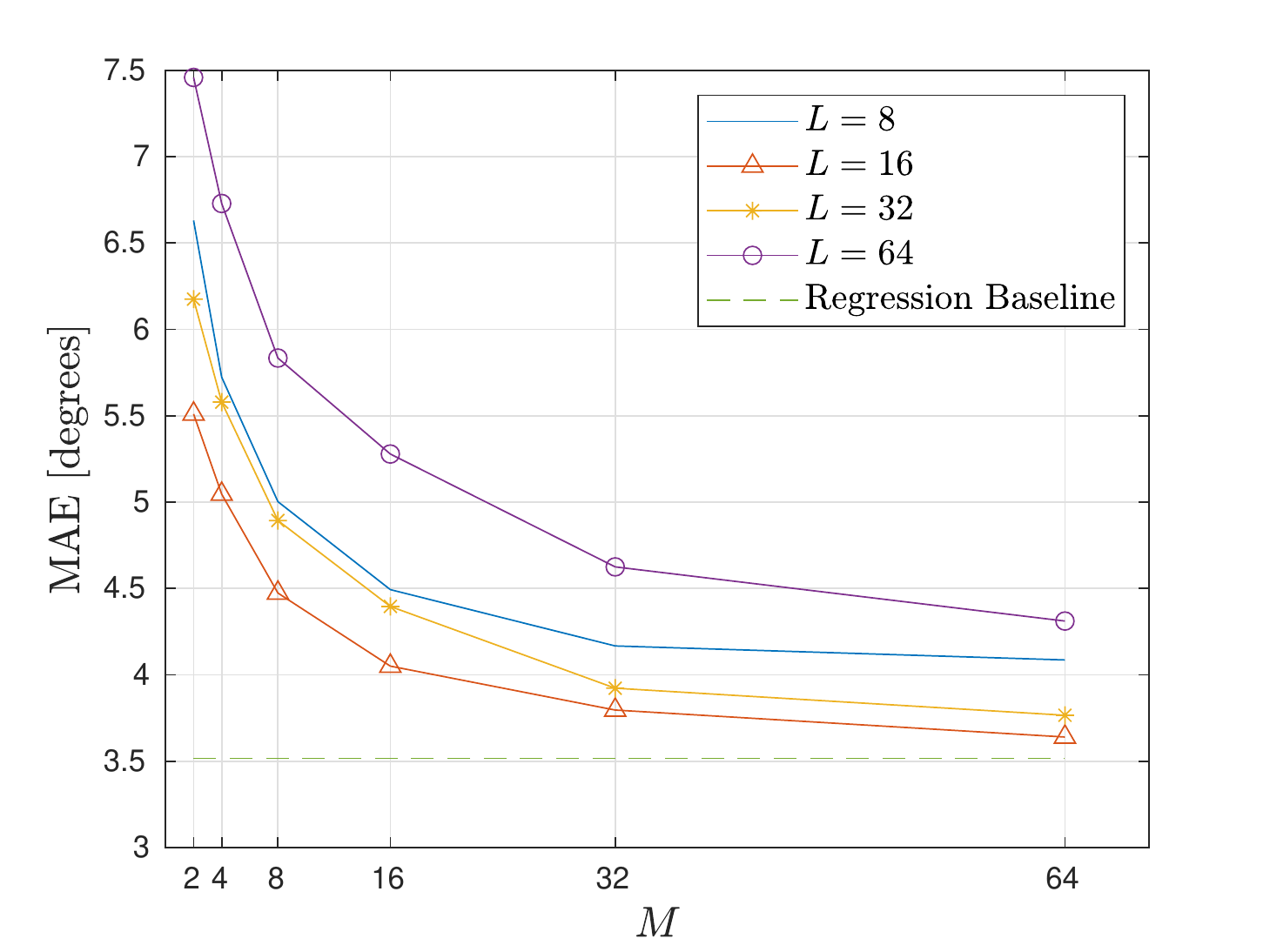}
  \caption{The MAE on the rotated MNIST datasets using the randomized bins method for different combinations of $L$ and $M$}
\label{LvsM}
\end{figure}

The results of the experiment is presented in Figure \ref{LvsM}. Here we clearly see the ensemble-like effect of our method, where the error decreases as the number of softmax heads $M$ is increased. This is expected from the error decomposition in equation (\ref{amb}). Additionally, we note that increasing the number of output units $L$, which leads to a decreased discretization error caused by the binning of the target, does not necessarily imply a decrease in prediction error. In this experiment, $L = 16$ yielded the smallest MAE for all values of $M$. This agrees with previous findings, namely that too few bins leads to a large discretization error, but too many can lead to poor convergence \cite{diaz2019soft, fu2018deep}. In general, the optimal number of bins depends on the specific task and finding it therefore requires an extensive parameter search.

\section{Experiments}
We compare our method with direct regression and classification baselines, as well as current state of the art methods on three challenging datasets. In order to make fair comparisons of the methods, we use the same baseline architecture for all experiments. More specifically, we use a pre-trained ResNet50 \cite{he2016deep} and replace the final fully connected layer with one fully connected layer of size 2048, a ReLU activation, and then a method-specific layer. For the direct regression approach (referred to as "Direct"), we use a fully connected layer of size 1 with a linear activation and train it by minimizing the MSE. For the RvC approach, we use one fully connected layer of size $K$ with a single softmax activation and train it using the cross-entropy loss function. For our own method we use $M$ fully connected layers of size $L$, with an individual softmax activation function on each layer, and train it by minimizing the sum of the individual cross-entropies for each discretization as in equation (\ref{ce}).

For all datasets, we train the network for 30 epochs using the ADAM optimizer \cite{kingma2014adam} with a mini-batch size of 32, a learning rate of 0.0005 that is decreased by a factor of 0.1 every 10th epoch, and an $L^2$-regularization factor of 0.001 on the weights. For data augmentation, we use random horizontal flipping of the images and apply a uniformly distributed random translation and scaling between [-20, 20] pixels and [0.7, 1.4] respectively. All results are averaged over 10 trials with different random initializations of the last fully connected layers. The experiments were implemented in Matlab and the network training was done using an NVIDIA Titan V graphics card. The source code for our experiments is available at \url{https://github.com/axeber01/dold}.

\subsection{Age Estimation}

\begin{figure}[t]
  \centering
  \includegraphics[width=\linewidth, trim=1cm 1cm 1cm 4cm,clip]{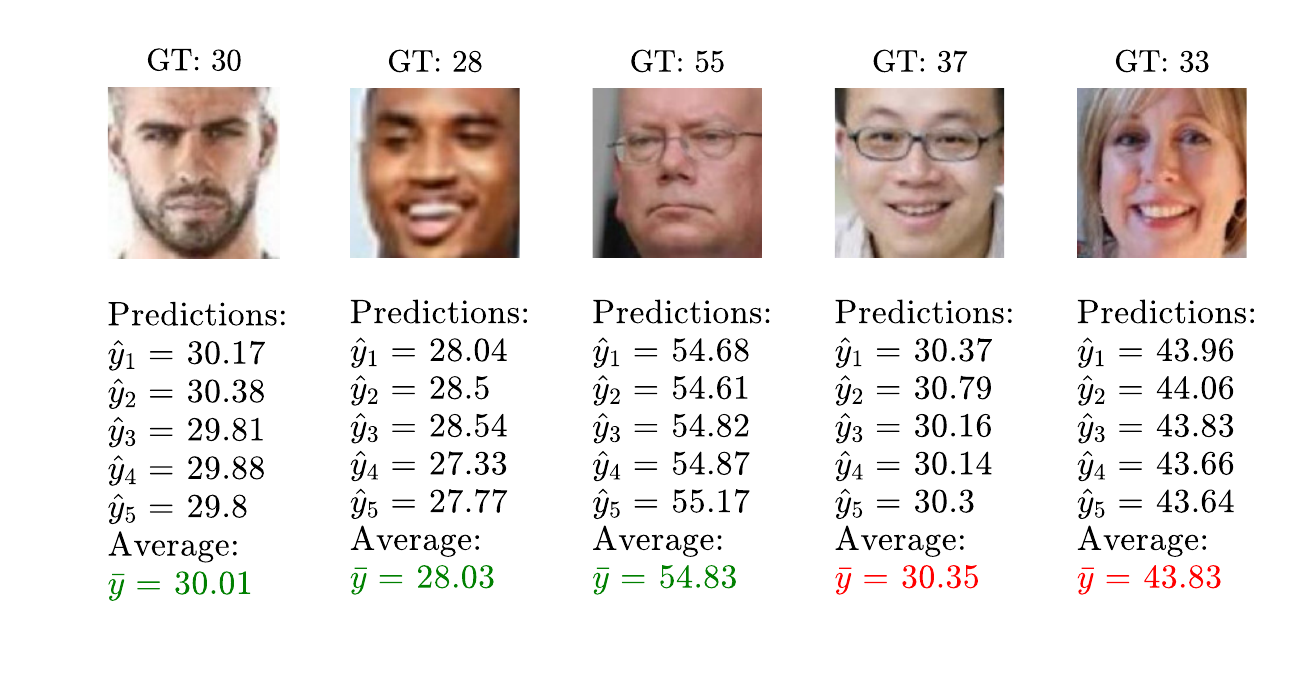}
  \caption{A subset of the images in UTKFace dataset \cite{zhifei2017cvpr} with ground truth (GT) labels and predictions using the equal width method.}
\label{utk_preview}
\end{figure}

For age estimation, we test our method on the UTKFace dataset \cite{zhifei2017cvpr}, which provides a large collection of images with human subjects labeled with ground truth ages. More specifically, we use the same train-test split of the data as in \cite{cao2019rank} and \cite{gustafsson2019dctd}, where the ages are in the set of integers $t_n \in [21, 60]$. The subset contains 13,144 training images and 3,287 test images. Furthermore, the images have been cropped such that only the faces of the subjects are visible, as is shown in the examples in Figure \ref{utk_preview}. 

For the RvC baseline, we simply use one class per age group in years, such that $\mathcal{C} = \{21, 22, ..., 60\}$. We compare the baseline with two implementations of the proposed method: equal-width overlapping bins and randomized bins. For the equal width approach we let $L = 8$ and $M = 5$, such that $LM = 40$, which means that the network complexity similar to the complexity of the RvC baseline. In practice this means that the first output head will classify images into age categories of $\mathcal{D}^1 = $\{21---25, 26---30, ..., 56---60\}, while the second output head will have $\mathcal{D}^2 = $\{21, 22---26, 27---31, ..., 57---60\} and so on.

For the randomized bins approach, we let $L = 10$ and $M = 30$ and draw a new sample of randomized bins for each run. This approach leads to a increased number of output heads compared to the RvC baseline, but the increase in model complexity is still small in relation to the size of the CNN backbone.

The results are shown and compared to current state-of-the art methods in Table 1, where we have evaluated the mean average error (MAE) of the different methods on the test set. The results are averaged over the trials with different random seeds and presented with the corresponding standard deviation. Of the two baselines, the direct method performs best. Since this is perhaps the most natural design choice, it should not come as a surprise, although other papers have reported contrary results on age estimation tasks \cite{rothe2015dex}. Both the equal width and randomized bins approach improve over the RvC baseline significantly, and they yield a small improvement over the direct method, which show that for this task, our proposed methods are more effective than the baseline methods. Furthermore, we reduce the average error compared to current state of the art by 2\%. We hypothesize that this error reduction is due to the diverse representation of the target values, which also incorporates the ordinal relationship between the age categories.

\begin{table*}[t]
\vspace{0.1in}
\renewcommand{\arraystretch}{1.3}
\caption{Mean average error in years for the different methods on the UTKFace \cite{zhifei2017cvpr} test set.}
\label{utkresults}
\centering
\begin{tabular}{c c c c c c c}
\hline
Method & CORAL \cite{cao2019rank}   & DCTD \cite{gustafsson2019dctd} & Direct (Ours) & RvC (Ours) & Equal Width (Ours) & Randomized Bins (Ours)\\
\hline
MAE & $5.47 \pm 0.01$ & $4.65 \pm 0.02$      & $4.60 \pm 0.02$ & $4.71 \pm 0.03$            & $4.58 \pm 0.03$ & \textbf{4.55 $\pm$ 0.04}             \\
\hline
\end{tabular}
\end{table*}

\subsection{Head Pose Estimation}
The BIWI dataset \cite{eth_biwi_00839} consists of 24 video sequences of 20 subjects recorded in a controlled environment and each frame is labeled with the corresponding head pose of the subject. We use the train-test split defined as protocol 2 in \cite{yang2019fsa}, where 16 videos are used for training and 8 for testing. In total, the training set consists of 10,063 images and the test set of 5,065 images. The pose is represented using the yaw, pitch and roll angles of the head, where each angle is approximately distributed in the range $t_n \in [-75 \degree, 75 \degree]$.

Following \cite{gustafsson2019dctd}, we use the same approach as for head pose estimation, but with small modifications needed to get a three dimensional output from the network. For the direct regression approach, we simply replace the last fully connected layer of size 1 by three fully connected layers of the same size. For the RvC, we use three chains of fully connected layers at the end, one for each angle, with a corresponding softmax head. We discretize each angle using 1 degree per bin, such that $\mathcal{C} = \{-75, -74, ..., 75\}$.

For the equal width approach, we let $M = 3$ and $L = 50$, such that $LM = 150$. Again, this gives a similar network complexity as the RvC baseline. Hence each softmax head has 3 degrees per bin, i.e. $\mathcal{D}^1 = $\{(-75) --- (-73), ..., 72 --- 75\}, $\mathcal{D}^2 = $\{-75, (-74) --- (-71), ..., 73 --- 75 \} and similarly for $\mathcal{D}^3$. For the randomized bins approach, we let $L = 20$ and $M = 30$ and make a new sample of randomized bins for each run. 

The results for the BIWI dataset are shown in Table \ref{biwiresults}, where we have evaluated the MAE for the three different angles for each method. On average, direct regression performs best, while the randomized bins method is best at predicting the yaw angle. However, both equal width and randomized bins outperform the standard RvC approach. Additionally, our direct method reduces the current state of the art average error by 16 \%, which shows that a carefully tuned regression baseline can outperform more sophisticated methods on this problem. 

\begin{figure}[t]
  \centering
  \includegraphics[width=\linewidth, trim=1cm 3cm 1cm 3cm,clip]{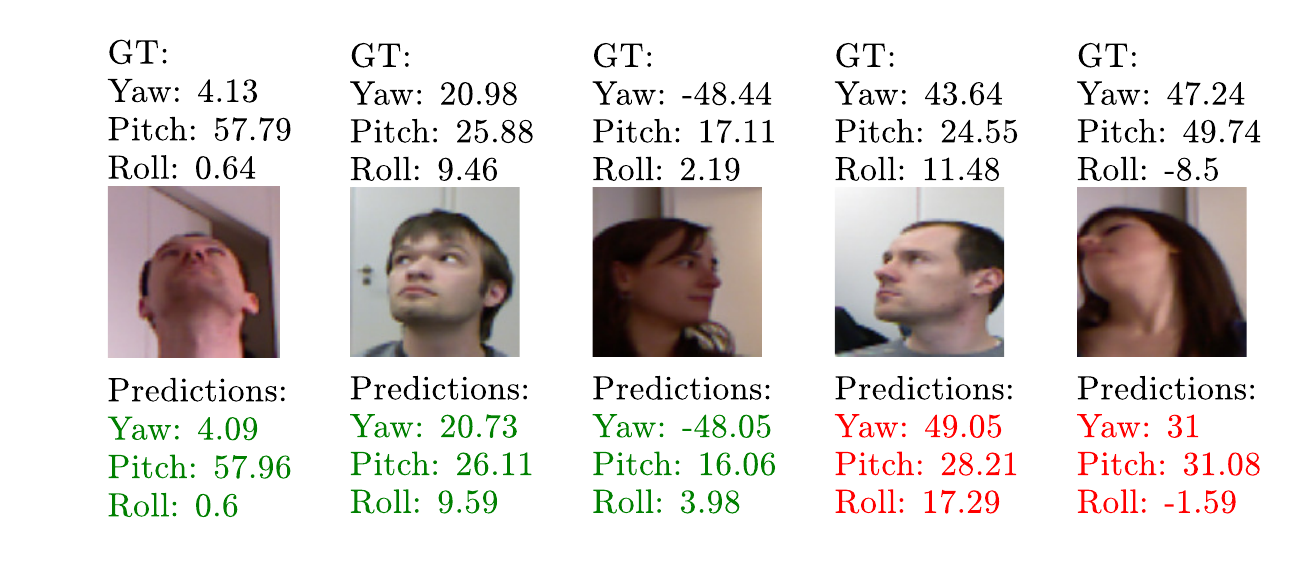}
  \caption{A subset of the images in BIWI dataset \cite{eth_biwi_00839} with ground truth (GT) labels and predictions using the equal width method.}
\label{biwi_preview}
\end{figure}

\begin{table*}[t]
\vspace{0.1in}
\renewcommand{\arraystretch}{1.3}
\caption{Mean average error in degrees for the different methods on the Biwi \cite{eth_biwi_00839} test set.}
\label{biwiresults}
\centering
\begin{tabular}{c c c c c c c}
\hline
Method & Yang et al. \cite{yang2019fsa}   & DCTD \cite{gustafsson2019dctd} & Direct (Ours) & RvC (Ours) & Equal Width (Ours) & Randomized Bins (Ours)\\ \hline
Yaw & 2.89 & $2.67 \pm 0.08$ & 2.64 $\pm$ 0.16 & 2.85 $\pm$ 0.12 & 2.62 $\pm$ 0.11 & \textbf{2.52 $\pm$ 0.06}  \\
Pitch & 4.29 & $3.61 \pm 0.12$ & \textbf{2.75 $\pm$ 0.05} & 3.22 $\pm$ 0.09 & 3.12 $\pm$ 0.08 & 3.01 $\pm$ 0.10 \\
Roll & 3.60 & $2.75 \pm 0.10$ & \textbf{2.24 $\pm$ 0.07} & 2.48 $\pm$ 0.08 & 2.33 $\pm$ 0.09 & 2.34 $\pm$ 0.10 \\
\hline
Average & $3.60$ & $3.01 \pm 0.07$ & \textbf{2.54 $\pm$ 0.05} & 2.85 $\pm$ 0.08 & 2.69 $\pm$ 0.04 & 2.63 $\pm$ 0.04 \\
\hline
\end{tabular}
\end{table*}

\subsection{Historical Image Dating}
In order to test our method on a small dataset with a small number of ordinal classes, we ran experiments on the Historical Color Images (HCI) dataset \cite{DBLP:conf/eccv/PalermoHE12}. The dataset consists of 1,375 color images from five decades, spanning from the 1930s to the 1970s. For evaluation, we use Monte Carlo random sampling with an 80/20 train-test split for each decade, drawn randomly at each iteration.

\begin{figure}[t]
  \centering
  \includegraphics[width=0.5\linewidth]{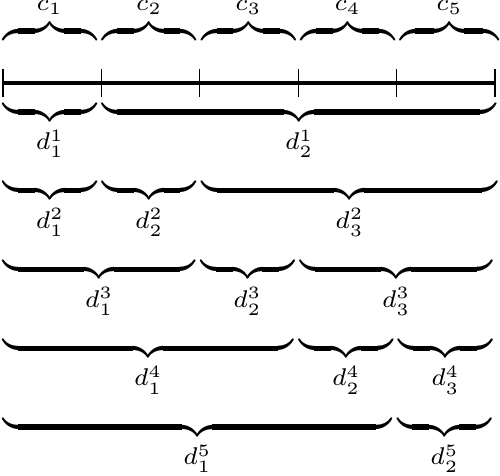}
  \caption{The sets of overlapping classes used for label diversity on the historical image dating task.}
  \label{hist_bins}
\end{figure}

For this dataset, the target value can then be considered as belonging to one of five ordinal classes $\mathcal{C} = \{c_1, c_2, c_3, c_4, c_5\}$, where each class corresponds to one of the five decades.  Likewise, the number of possibilities for creating our new sets of bins is limited, so the methods of overlapping and randomized bins are not suitable. We instead create 5 new sets $\{\mathcal{D}^m\}_{m=1}^5$ as shown in Figure \ref{hist_bins} and refer to this method simply as "Label Diversity". Although it is possible to define a distance metric between classes as the distance in decades, this is unsuitable, since the year 1939 is closer to 1940 than it is to 1949, but the distances in decades are the same. We therefore treat this as an ordinal classification problem and use the MAP estimate in equation (\ref{map}) for inference. 

\begin{figure}[t]
  \centering
  \includegraphics[width=\linewidth, trim=1cm 4cm 1cm 4cm,clip]{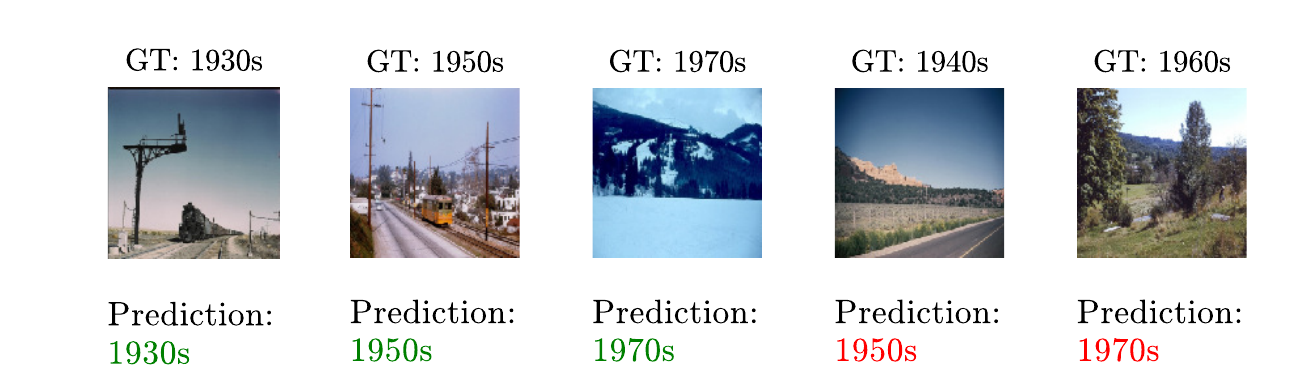}
  \caption{A subset of the images in HCI dataset \cite{DBLP:conf/eccv/PalermoHE12} with ground truth (GT) labels and predictions using the label diversity method.}
\label{hci_preview}
\end{figure}

We evaluate the methods in terms of classification accuracy and MAE. A sample of correct and incorrect predictions are shown in Figure \ref{hci_preview}. The results are shown in Table \ref{hciresults} and we conclude that using label diversity improves both accuracy and MAE over the regression baseline. Label diversity also decreases MAE compared to the RvC baseline, which we claim is due to the exploitation of the ranking between classes. Furthermore, our method improves the accuracy by one percentage point compared to the current state of the art method \cite{belharbi2019deep}.

\begin{table*}[t]
\renewcommand{\arraystretch}{1.3}
\caption{Accuracy and mean average error in decades for the different methods on the HCI dataset \cite{DBLP:conf/eccv/PalermoHE12}.}
\label{hciresults}
\centering
\begin{tabular}{c c c c c c c c}
\hline
Method & CNNOr \cite{liu2017deep}   & PN \cite{belharbi2019deep} & ELB \cite{belharbi2019deep} & Direct (Ours) & RvC (Ours) & Label Diversity (Ours)\\
\hline
Acc (\%) & 41.56 &  56.67 $\pm$ 2.30 & $54.90 \pm 2.40$ & $46.9 \pm 1.7$ & 57.1 $\pm$ 1.9 & \textbf{57.7 $\pm$ 2.0}    \\
MAE & 1.04 & $0.67 \pm 0.05$ & \textbf{0.63 $\pm$ 0.04} & $0.76 \pm 0.01$ & $0.72 \pm 0.04$ & $0.67 \pm 0.03$ \\ 
\hline
\end{tabular}
\end{table*}

\section{Conclusion}
In this work, we have shown that employing a series of different discrete representations of the target values, it is possible to improve the predictive performance of a deep neural network, compared to when using a single such representation. For some problems, it can also outperform direct regression. We note that our methods yield the strongest improvement compared to the regression baseline on historical image dating, where the target belongs to a small set of ordinal classes. For head pose estimation, where the target is continuous, the improvement was either small or negligible, but there is a significant improvement over the RvC baseline. For age estimation, where the target belongs to a large set of ordered classes, there is an improvement over both the regression and RvC method. Nevertheless, our method consistently improves over the RvC baseline for all methods. Hence our conclusion is that, if it is suitable to approach a regression problem via classification, then using several diverse representations can improve performance. 

This opens up a wide range of options when it comes to selecting the representations, since there are many ways to create different discrete binnings of a continuous target value. As we have also shown, the number of discretizations and the number of bins for each discretization will have an impact on the prediction performance, but since these choices also affect the training convergence, it is difficult to select them for a given problem without extensive parameter search. In future research we will continue to investigate these questions and how diversity in label representations can be exploited in other ways.

\section*{Acknowledgment}
This work was supported by the Wallenberg Artificial Intelligence, Autonomous Systems and Software Program (WASP), funded by Knut and Alice Wallenberg Foundation.

\bibliographystyle{IEEEtran}
\bibliography{references.bib}

\end{document}